# A Nonlinear PID-Enhanced Adaptive Latent Factor Analysis Model

Jinli Li, Ye Yuan

*Abstract*—High-dimensional and incomplete (HDI) data holds tremendous interactive information in various industrial applications. A latent factor (LF) model is remarkably effective in extracting valuable information from HDI data with stochastic gradient decent (SGD) algorithm. However, an SGD-based LFA model suffers from slow convergence since it only considers the current learning error. To address this critical issue, this paper proposes a Nonlinear PID-enhanced Adaptive Latent Factor (NPALF) model with two-fold ideas: 1) rebuilding the learning error via considering the past learning errors following the principle of a nonlinear PID controller; b) implementing all parameters adaptation effectively following the principle of a particle swarm optimization (PSO) algorithm. Experience results on four representative HDI datasets indicate that compared with five state-of-the-art LFA models, the NPALF model achieves better convergence rate and prediction accuracy for missing data of an HDI data.

## I. INTRODUCTION

With the development of big data, the internet is releasing lots of data all the time, e.g., some big-data-related applications hold numerous users, items and massive amounts of interactive information [1-12]. However, as the number of users and items consistent growth, the proportion of interactive information is shrinking, e.g., each user only touch a small part of item, but fail to contact with richer items [3, 4]. Thus, it is highly incomplete of resultant interactions in many applications. In consequence High-dimensional and incomplete (HDI) matrix commonly used to characterize interactive information with a high deficiency rate [5-7, 9].

Even though an HDI matrix exists a lot of blank terms, it still contains plentiful interactive results of users and items, which helps to calculate user preferences and attributes of items [5, 6, 14]. Hence, how to extract target information efficiently and accurately from known interactive data in HDI matrix becomes a focused issue and trend in research [7, 31, 33]. Much research indicated that latent factor analysis (LFA) model obtained great success in acquiring useful information from HDI data [33, 40, 41, 44]. LFA model maps both users and items into identical low-dimensional LF space, and then calculate its low-rank approximation based on known data in HDI. In general, LFA has become the mainstream model for explaining HDI matrix by feat of its scalability and efficiency [7, 31, 33, 34, 36, 39].

Stochastic gradient descent (SGD) method is commonly considered as one of the most suitable optimization algorithms for explaining LFA model [20, 43, 45]. It utilizes negative gradient information of each point for helping objection convergence. However, a standard SGD algorithm only takes current gradient information into consideration instead of all known historical information of this point. Thus, SGD-based LF model consumes multiple iterations to converge and costs more training time and resources on large-scale datasets [45]. Therefore, many researchers are devoted to investigating how to accelerate the training speed based on SGD-based LFA without sacrificing the accuracy [20]. To date, many studies have put forward a series of optimization algorithms based on SGD by incorporating past gradient information. For instance, a momentum-incorporated SGD algorithm, it considers both the past and present gradients to update the parameter [13, 28, 41]. A Nesterov's accelerated gradient (NAG)-incorporated SGD algorithm is a variant of momentum algorithm, but the gradient is estimated after the current velocity is applied [32]. In conclusion, above methods introduce history gradient information by adding momentum term.

It has been found that proportional-integral-derivative (PID) popularly employed in control field due to its simplicity, functionality and applicability [16-19]. It is a feedback-loop just with a single input, and it contains the proportion of current error, the integral of all history error, and the derivative of future trend (the difference between recent two errors) [22, 23]. Inspired by PID method, a PID-incorporated SGD algorithm is proposed to improve the convergence rate without accuracy loss [38]. Specifically, it combines the current, past and future trend of error to rebuild a new error, and then replace current error of SGD with rebuild error. However, the above model cannot deal with nonlinear problems well. Thus, a Non-linear PID-incorporated SGD-based LF (NSLF) algorithm is proposed to solve a complex nonlinear problem [37], However, there are many gain parameters and hyper-parameters like lambda and learning rate requires gird search, this issue significantly affects model performance when processing an industrial HDI matrix.

Aiming at addressing the above issue, this paper innovatively proposes a <u>N</u>onlinear <u>P</u>ID-enhanced <u>A</u>daptive <u>L</u>atent <u>F</u>actor (NPALF) algorithm with following two-fold ideas:

a) Replacing lambda times learning rate with a new constant $\varphi$, then using NPID rebuilding the instant error, which replaces the original instant error in SGD-based LF model with rebuilt error to achieve fast convergence rate and competitive prediction accuracy for missing data of an HDI matrix; and
b) Enabling $\varphi$ and gain parameters of NPID adaptation in formula, thereby improving the efficiency of LFA model significantly.

Main contributions of this work as shown:

a) **An NPALF-based LFA model.** Using NPID-based instant error control and PSO for adapting hyper-parameter and gain parameters of NPID, an NPALF-based LFA model achieves efficient convergence rate and reliable accuracy on HDI matrix.
b) **Detail algorithm design and analysis for an NPALF-based LFA model.** It explains that the NPALF algorithm makes LFA model obtain better results in terms of calculation and storage.

Experimental results are conducted on four HDI matrices generated by real applications to prove that an NPALF-based LFA model's performance of convergence and accuracy is better than other popular SGD-based LFA models. The rest part of this paper is organized as follows. Section II describes the preliminaries, Section III accounts for NPALF-based LFA model, Section IV reports the experiment and comparison of results, and finally, there are conclusions about this study in Section V.

## II. PRELIMINARIES

### A. Problem Statement

***Definition* 1:** Given $M$ and $N$ denote the user and item sets respectively, $R^{|M|\times|N|}$ is rating matrix whose each value $r_{m,n}$ is user $m$'s preference on item $n$, $\Lambda$ and $\Gamma$ denote the known and unknown entity sets of $R$, then $R$ is an HDI matrix if $|\Lambda|\ll|\Gamma|$.

Given $R$, the definition of an LFA model:

***Definition* 2:** Given $R$, an LFA model is its low-rank estimation $\hat{R}=XY^T$ built on $R_\Lambda$, where $X^{|M|\times|f|}$ and $Y^{|N|\times|f|}$ are the LF matrices corresponding to $M$ and $N$, and $f\ll\min\{|M|, |N|\}$ denotes the rank of $\hat{R}$.

Obviously, matrices $X$ and $Y$ are importance for an LFA model. In order to achieve them, an LFA model builds a learning objective to measure the difference between each node in $R_\Lambda$ and corresponding node in $\hat{R}$ [25-29]. With Euclidean distance and regularization [7-9], it is formulated as:

$$\varepsilon = \frac{1}{2}\sum_{r_{m,n}\in\Lambda}\left(\left(r_{m,n}-\hat{r}_{m,n}\right)^2+\lambda\|x_m\|_2^2+\lambda\|y_n\|_2^2\right), \tag{1}$$

where regularization constant $\lambda$ controls the regularization effect [21], and $\|\cdot\|_2$ computes the $l_2$ norm of a vector.

### B. An SGD-based LFA model

As indicated by prior studies [20, 45], SGD algorithm relies it's highly efficient when performing LFA of HDI matrices. Thus, with SGD method, the objective function (1) is shown as:

$$\underset{X,Y}{\arg\min}\,\varepsilon \overset{SGD}{\Rightarrow} \forall r_{m,n}\in\Lambda:\begin{cases}x_m \leftarrow x_m - \eta\dfrac{\partial\varepsilon_{m,n}}{\partial x_m},\\ y_n \leftarrow y_n - \eta\dfrac{\partial\varepsilon_{m,n}}{\partial y_n};\end{cases} \tag{2}$$

where $\eta$ denotes the learning rate; and the instant objective $\varepsilon_{m,n}$ is given as:

$$\varepsilon_{m,n} = \left(r_{m,n}-\langle x_m, y_n\rangle\right)^2+\lambda\|x_m\|_2^2+\lambda\|y_n\|_2^2. \tag{3}$$

By making $err_{m,n}=r_{m,n}-\langle x_m, y_n\rangle$, and folding (2)-(3), we achieve an SGD-based LFA model as follow:

$$\underset{X,Y}{\arg\min}\,\varepsilon \overset{SGD}{\Rightarrow} \forall r_{m,n}\in\Lambda:\begin{cases}x_m \leftarrow x_m + \eta\cdot\left(e_{m,n}\cdot y_n - \lambda\cdot x_m\right),\\ y_n \leftarrow y_n + \eta\cdot\left(e_{m,n}\cdot x_m - \lambda\cdot y_n\right).\end{cases} \tag{4}$$

### C. An NPID Controller

The PID controller exploits the present, past and future information of prediction error to calculate a correction and applies it to control a feedback system [30, 35]. However, a standard discrete PID controller adopts constants as gain parameters of PID, so that it cannot adjust the control effect more flexibly in real time according to the current feedback result. Therefore, NPID is proposed into an SGD algorithm [47-49]. Unlike traditional PID, gain parameters are nonlinear functions with the instant error as the independent variable. Hence, NPID can adjust the control effect in real time according to the instant error, i.e., $K_p$, $K_i$ and $K_d$ in NPID is shown as:

$$\begin{cases} K_p(e^t) = K_{p1} + K_{p2}(1-\mathrm{sech}(K_{p3}e^t)), \\ K_i(e^t) = K_{i1}\mathrm{sech}(K_{i2}e^t), \\ K_d(e^t) = K_{d1} + K_{d2}/(1+K_{d3}\exp(K_{d4}e^t)); \end{cases} \quad (5)$$

where $e^{(t)}$ represent the $t$-th instant error, $K_{p_1} \sim K_{p_3}, K_{i_1} \sim K_{i_2}$ and $K_{d_1} \sim K_{d_4}$ are proportional, integral and derivative parameters, nonlinear function $\mathrm{sech}(x)$ is expressed as $2/(\exp(x)+\exp(-x))$, respectively. With function (5), the NPID expression expands as follow:

$$\begin{aligned}\tilde{e}^\tau = {} & (K_{p1}+K_{p2}(1-\mathrm{sech}(K_{p3}e^\tau)))e^\tau \\ & + K_{i1}\mathrm{sech}(K_{i2}e^\tau)\sum_{k=0}^{\tau}e^\tau \\ & + (K_{d1}+K_{d2}/(1+K_{d3}\exp(K_{d4}e^\tau)))(e^\tau-e^{\tau-1}).\end{aligned} \quad (6)$$

*D. A PSO Algorithm*

Note that a standard PSO algorithm originated from the foraging behavior of bird flocks [40, 42, 46]. It is a population-based stochastic optimization technique. In general, PSO builds a swarm with $J$ particles, and those particles search for the optimal solution in a $D$-dimensional space. Each particle $j$ has two intrinsic properties: velocity $v_j$ and position vectors $s_j$. In detail, $v_j$ and $s_j$ at the $t$-th iteration are vectors $v_j^{(t)}=\left[v_{j,1}^{(t)},v_{j,2}^{(t)},\ldots,v_{j,d}^{(t)},\ldots,v_{j,D}^{(t)}\right]$, $s_j^{(t)}=\left[s_{j,1}^{(t)},s_{j,2}^{(t)},\ldots,s_{j,d}^{(t)},\ldots,s_{j,D}^{(t)}\right]$ where $1\leq d\leq D$. Thus, its evolution scheme of the $j$-th particle at the $t$-th iteration is as follows:

$$\forall j \in \{1,\ldots,J\}: \begin{cases} v_j^{(t)} = wv_j^{(t-1)} + c_1r_1\left(b_j^{(t-1)}-s_j^{(t-1)}\right) + c_2r_2\left(g^{(t-1)}-s_j^{(t-1)}\right), \\ s_j^{(t)} = s_j^{(t-1)} + v_j^{(t)}; \end{cases} \quad (7)$$

where $b_j^{(t-1)}$ and $g^{(t-1)}$ denote the $j$-th particle's local optimal position and the global position after the $(t-1)$-th iteration, $s_j^{(t-1)}$ and $v_j^{(t-1)}$ denotes the state of position and velocity after the $(t-1)$-th iteration, respectively.

III. METHODS

*A. NPID-based Error Refinement*

A NPID controller will make calculate a correction based on the instant error $e^{(t)}$ between the prediction value and the true value. Starting from our problem, we handle each single instance $r_{m,n}\in\Lambda$ with such refinement, and consider each iteration is a time point in NPID controller given in (8). Actually, we implement a refinement sequence of $\Lambda$ by execute NPID controller of each one. Following such principle, the SGD-based learning principle (4) as follow:

$$\forall r_{m,n} \in \Lambda : \begin{cases} x_m \leftarrow x_m + \eta\cdot\left(e_{m,n}^{(t)}\cdot y_n - \lambda\cdot x_m\right), \\ y_n \leftarrow y_n + \eta\cdot\left(e_{m,n}^{(t)}\cdot x_m - \lambda\cdot y_n\right); \end{cases} \quad (8)$$

where $e_{m,n}^{(t)}$ represent the estimation error of the instance $r_{m,n}$ at the $t$-th training iteration, also the $t$-th time point for an NPID controller. Thus, the NPID based error refining rule is shown as below:

$$\begin{aligned}\tilde{e}_{m,n}^t = {} & (K_{p1}+K_{p2}(1-\mathrm{sech}(K_{p3}e_{m,n}^t)))e_{m,n}^t \\ & + K_{i1}\mathrm{sech}(K_{i2}e_{m,n}^t)\sum_{k=0}^{\tau}e_{m,n}^k \\ & + (K_{d1}+K_{d2}/(1+K_{d3}\exp(K_{d4}e_{m,n}^t)))(e_{m,n}^t-e_{m,n}^{t-1}).\end{aligned} \quad (9)$$

Formula (9) can be interpreted as follows:

a) $e_{m,n}^{(t)}$ represents the current training residual of an LFA;

b) $\sum_{k=0}^{\tau}e_{m,n}^k$ represents the history residual of $r_{m,n}$ from the beginning till current;

c) $e_{m,n}^{t} - e_{m,n}^{t-1}$ represents future trend of $r_{m,n}$, it helps avoid overshooting.

By combining (8)-(9), we obtain the NPID incorporated SGD-based learning scheme for an LFA model as (10):

So far, an NPID-based LFA model is achieved. However, formula (10) has a large number of gain parameters and hyper-parameters to adjust, which is time-consuming. In order to solve this problem, section III-B introduce PSO algorithm for realizing all parameters adaptation in (10).

$$\arg\min_{X,Y} \varepsilon \xrightarrow{NPID-SGD}$$

During the $\iota$-th training interation, $\forall r_{m,n} \in \Lambda$:

$$\begin{aligned}
\tilde{e}_{m,n}^{(t)} &= \left(K_{p1} + K_{p2}\left(1-\operatorname{sech}\left(K_{p3}e_{m,n}^{(t)}\right)\right)\right)e_{m,n}^{(t)} \\
&+ K_{i1}\operatorname{sech}\left(K_{i2}e_{m,n}^{(t)}\right)\sum_{k=0}^{\tau}e_{m,n}^{(k)} \\
&+ \left(K_{d1} + K_{d2}\Big/\left(1+K_{d3}\exp\left(K_{d4}e_{m,n}^{(t)}\right)\right)\right)\left(e_{m,n}^{(t)} - e_{m,n}^{(t-1)}\right), \\
x_m &\leftarrow x_m + \eta\cdot\left(\tilde{e}_{m,n}^{(\tau)}\cdot y_n - \lambda\cdot x_m\right), \\
y_n &\leftarrow y_n + \eta\cdot\left(\tilde{e}_{m,n}^{(\tau)}\cdot x_m - \lambda\cdot y_n\right).
\end{aligned} \quad (10)$$

## B. Self-adaptation of All Parameters

NPID-incorporated SGD-based LFA model (10) has nine gain parameters and two hyper-parameters need to adjust, which is time-consuming. This will limit the use of NPID-incorporated SGD learning scheme in large industrial application. Therefore, we introduce the PSO algorithm to NPID-incorporated SGD method for implementing the self-adaptation of all parameters.

We combine regularization coefficient $\lambda$ times learning rate $\eta$ into a new constant term $\varphi$, then, we expand $\tilde{e}_{m,n}^{(t)}$ in (10) with NPID method, and multiply $\eta$ into NPID method, equation (10) is modified as follows:

$$\begin{cases}
x_m \leftarrow (1-\varphi)\cdot x_m + \begin{pmatrix} \left(\eta\cdot K_{p1} + \eta\cdot K_{p2}\left(1-\operatorname{sech}\left(K_{p3}e_{m,n}^{(t)}\right)\right)\right)e_{m,n}^{(t)} \\ +\eta\cdot K_{i1}\operatorname{sech}\left(K_{i2}e_{m,n}^{(t)}\right)\sum_{k=0}^{t}e_{m,n}^{(t)} \\ +\left(\eta\cdot K_{d1} + \eta\cdot K_{d2}\Big/\left(1+K_{d3}\exp\left(K_{d4}e_{m,n}^{(t)}\right)\right)\right)\left(e_{m,n}^{(t)} - e_{m,n}^{(t-1)}\right) \end{pmatrix} y_n, \\
y_n \leftarrow (1-\varphi)\cdot y_n + \begin{pmatrix} \left(\eta\cdot K_{p1} + \eta\cdot K_{p2}\left(1-\operatorname{sech}\left(K_{p3}e_{m,n}^{(t)}\right)\right)\right)e_{m,n}^{(t)} \\ +\eta\cdot K_{i1}\operatorname{sech}\left(K_{i2}e_{m,n}^{(t)}\right)\sum_{k=0}^{t}e_{m,n}^{(t)} \\ +\left(\eta\cdot K_{d1} + \eta\cdot K_{d2}\Big/\left(1+K_{d3}\exp\left(K_{d4}e_{m,n}^{(t)}\right)\right)\right)\left(e_{m,n}^{(t)} - e_{m,n}^{(t-1)}\right) \end{pmatrix} x_m.
\end{cases} \quad (11)$$

After that, we multiply the learning rate into the corresponding term. Meanwhile, we represent the parameters that need to be adapted by a set of unified symbols as follow:

$$\begin{cases}
x_m \leftarrow (1-\tilde{K}_{\varphi})\cdot x_m + \begin{pmatrix} \left(\tilde{K}_{P1} + \tilde{K}_{P2}\left(1-\operatorname{sech}\left(\tilde{K}_{p3}e_{m,n}^{(t)}\right)\right)\right)e_{m,n}^{(t)} \\ +\tilde{K}_{i1}\operatorname{sech}\left(\tilde{K}_{i2}e_{m,n}^{(t)}\right)\sum_{k=0}^{t}e_{m,n}^{(t)} \\ +\left(\tilde{K}_{d1} + \tilde{K}_{d2}\Big/\left(1+\tilde{K}_{d3}\exp\left(\tilde{K}_{d4}e_{m,n}^{(t)}\right)\right)\right)\left(e_{m,n}^{(t)} - e_{m,n}^{(t-1)}\right) \end{pmatrix} y_n, \\
y_n \leftarrow (1-\tilde{K}_{\varphi})\cdot y_n + \begin{pmatrix} \left(\tilde{K}_{p1} + \tilde{K}_{p2}\left(1-\operatorname{sech}\left(\tilde{K}_{p3}e_{m,n}^{(t)}\right)\right)\right)e_{m,n}^{(t)} \\ +\tilde{K}_{i1}\operatorname{sech}\left(\tilde{K}_{i2}e_{m,n}^{(t)}\right)\sum_{k=0}^{t}e_{m,n}^{(t)} \\ +\left(\tilde{K}_{d1} + \tilde{K}_{d2}\Big/\left(1+\tilde{K}_{d3}\exp\left(\tilde{K}_{d4}e_{m,n}^{(t)}\right)\right)\right)\left(e_{m,n}^{(t)} - e_{m,n}^{(t-1)}\right) \end{pmatrix} x_m.
\end{cases} \quad (12)$$

Formula (12) gives the complete NPID model with constant $\varphi$. Thus, the NPID-incorporated SGD-based learning scheme depends on ten parameters, e.g. $\tilde{K}_\varphi, \tilde{K}_{p1}, \tilde{K}_{p2}, \tilde{K}_{p3}, \tilde{K}_{i1}, \tilde{K}_{i2}, \tilde{K}_{d1}, \tilde{K}_{d2}, \tilde{K}_{d3}, \tilde{K}_{d4}$. Thus, we construct a swarm with $J$ particles, which search for the optimal solution in $D$-dimensional space ($D=10$). And $j$-th particle maintains a set of parameters in an NPID controller applied to the same group of LFs. Then, the $j$-th particle's position and velocity vectors are given as follow:

$$\begin{cases} s_j = \left[s_{j,1}, s_{j,2}, s_{j,3}, s_{j,4}, s_{j,5}, s_{j,6}, s_{j,7}, s_{j,8}, s_{j,9}, s_{j,10}\right] \\ \quad = \begin{bmatrix} \tilde{K}_{j\varphi}, \tilde{K}_{jp_1}, \tilde{K}_{jp_2}, \tilde{K}_{jp_3}, \tilde{K}_{ji_1}, \\ \tilde{K}_{ji_2}, \tilde{K}_{jd_1}, \tilde{K}_{jd_2}, \tilde{K}_{jd_3}, \tilde{K}_{jd_4} \end{bmatrix}, \\ v_j = \left[v_{j,1}, v_{j,2}, v_{j,3}, v_{j,4}, v_{j,5}, v_{j,6}, v_{j,7}, v_{j,8}, v_{j,9}, v_{j,10}\right] \\ \quad = \begin{bmatrix} v_{jK_\varphi}, v_{jK_{p_1}}, v_{\mathcal{O}_{p_2}}, v_{\mathcal{O}_{p_3}}, v_{\mathcal{O}_{i_1}}, \\ v_{\mathcal{O}_{i_2}}, v_{\mathcal{O}_{d_1}}, v_{\mathcal{O}_{d_2}}, v_{\mathcal{O}_{d_3}}, v_{\mathcal{O}_{d_4}} \end{bmatrix}. \end{cases} \quad (13)$$

By combining (7) with (13), we obtain the evolution process of parameters with PSO algorithm. As shown in (7), PSO changes the state of $j$-th particles based on $b_j$ and $g$ in each iteration, whose update process is as follow:

$$\forall j \in \{1,\ldots,J\}: \begin{cases} b_j^{(t-1)} = \begin{cases} b_j^{(t-1)}, & F\left(b_j^{(t-1)}\right) \le F\left(s_j^{(t)}\right), \\ s_j^{(t)}, & F\left(b_j^{(t-1)}\right) > F\left(s_j^{(t)}\right); \end{cases} \\ g^{(t)} = \begin{cases} g^{(t-1)}, & F\left(g^{(t-1)}\right) \le F\left(s_j^{(t)}\right), \\ s_j^{(t)}, & F\left(g^{(t-1)}\right) > F\left(s_j^{(t)}\right). \end{cases} \end{cases} \quad (14)$$

In order to fitting the known set $\Lambda$ better, we adopt following two fitness function $F(\cdot)$ for the $j$-th particle:

$$\begin{cases} F_1(j) = \sqrt{\left(\sum_{r_{m,n} \in \Omega} \left(r_{m,n} - \hat{r}_{(j)m,n}\right)^2\right) \Big/ |\Omega|}, \\ F_2(j) = \left(\sum_{r_{m,n} \in \Omega} \left|r_{m,n} - \hat{r}_{(j)m,n}\right|_{abs}\right) \Big/ |\Omega|; \end{cases} \quad (15)$$

where $|\ |_{abs}$ represent the absolute value of the data, $\Omega$ represents the validation data, $\hat{r}_{(j)m,n}$ represents the prediction value of $r_{m,n}$. After that, we set a range for position and velocity of each particle to ensure them is constrained in a certain range.

$$\begin{cases} s_j^{(t)} = \min\left(\breve{s}, \max\left(\hat{s}, s_j^{(t)}\right)\right), \\ v_j^{(t)} = \min\left(\breve{v}, \max\left(\hat{v}, v_j^{(t)}\right)\right); \end{cases}$$
$$\Downarrow \forall d \in \{1,\ldots,D\} \quad (16)$$
$$\begin{cases} s_{j,d}^{(t)} = \min\left(\breve{s}_d, \max\left(\hat{s}_d, s_{j,d}^{(t)}\right)\right), \\ v_{j,d}^{(t)} = \min\left(\breve{v}_d, \max\left(\hat{v}_d, v_{j,d}^{(t)}\right)\right); \end{cases}$$

where $\breve{v}$ and $\hat{v}$ denote upper and lower bounds of the particle's velocity, $\breve{s}$ and $\hat{s}$ denote upper and lower bounds of the particle's position. In general, we commonly have $\hat{v} = 0.01 \times 2^{-\bar{s}} - 2^{-\bar{s}}$, and $\breve{v} = -\hat{v}$.

In addition, for $\forall j \in \{1,\ldots,J\}$, $s_j$ is linked with the same group of LF matrices, i.e., $X$ and $Y$. Thus, each iteration contains $J$ sub-iterations. And in $j$-th sub-iteration of the $t$-th iteration, $X$ and $Y$ are trained as:

$$\begin{cases} x_m \leftarrow \left(1-\tilde{K}_{(j)\varphi}^{(t-1)}\right)\cdot x_m + \begin{pmatrix} \left(\tilde{K}_{(j)p1}^{(t-1)} + \tilde{K}_{(j)p2}^{(t-1)}\left(1-\operatorname{sech}\left(\tilde{K}_{(j)p3}^{(t-1)}e_{m,n}^{(t)}\right)\right)\right)e_{m,n}^{(t)} \\ +\tilde{K}_{(j)i1}^{(t-1)}\operatorname{sech}\left(\tilde{K}_{(j)i2}^{(t-1)}e_{m,n}^{(t)}\right)\sum_{k=0}^{t} e_{m,n}^{(t)} \\ +\left(\tilde{K}_{(j)d1}^{(t-1)} + \tilde{K}_{(j)d2}^{(t-1)}\Big/\left(1+\tilde{K}_{(j)d3}^{(t-1)}\exp\left(\tilde{K}_{(j)d4}^{(t-1)}e_{m,n}^{(t)}\right)\right)\right)\left(e_{m,n}^{(t)}-e_{m,n}^{(t-1)}\right) \end{pmatrix} y_n, \\ y_n \leftarrow \left(1-\tilde{K}_{(j)\varphi}^{(t-1)}\right)\cdot y_n + \begin{pmatrix} \left(\tilde{K}_{(j)p1}^{(t-1)} + \tilde{K}_{(j)p2}^{(t-1)}\left(1-\operatorname{sech}\left(\tilde{K}_{(j)p3}^{(t-1)}e_{m,n}^{(t)}\right)\right)\right)e_{m,n}^{(t)} \\ +\tilde{K}_{(j)i1}^{(t-1)}\operatorname{sech}\left(\tilde{K}_{(j)i2}^{(t-1)}e_{m,n}^{(t)}\right)\sum_{k=0}^{t} e_{m,n}^{(t)} \\ +\left(\tilde{K}_{(j)d1}^{(t-1)} + \tilde{K}_{(j)d2}^{(t-1)}\Big/\left(1+\tilde{K}_{(j)d3}^{(t-1)}\exp\left(\tilde{K}_{(j)d4}^{(t-1)}e_{m,n}^{(t)}\right)\right)\right)\left(e_{m,n}^{(t)}-e_{m,n}^{(t-1)}\right) \end{pmatrix} x_m. \end{cases} \quad (17)$$

where the subscript ($j$) on $X$ and $Y$ represents their current states are linked with the $j$-th particle. According to formula (8)-(17), we obtain an NPALF-based LFA model.

## IV. EXPERIMENTAL RESULTS AND ANALYSIS

### A. General Settings

**Evaluation Protocol**. For industrial applications, the major issue is to recover the full connections among involved entities of HDI matrix. Thus, the root mean squared error (RMSE) is widely-adopted as evaluation metrics, the lower RMSE represent higher prediction accuracy:

$$RMSE = \sqrt{\left(\sum_{r_{u,i}\in\Phi} \left(r_{u,i}-\tilde{r}_{u,i}\right)^2\right)\Big/|\Phi|},$$

where $\hat{r}_{m,n}$ represents the prediction to $r_{m,n}$ generated on test data, $\Phi$ represents the testing set, $|\cdot|$ calculates the cardinality of an enclosed set. At the same time, we record the convergent rounds and time-consuming in each descend. Note that all experiments are carried out on the same PC with a 3.2 GHz i5 CPU and 16 GB RAM. All tested models are implemented in JAVA SE 7U60.

**Datasets.** The detail of Four HDI matrices adopted in our experiments is shown as below:

TABLE I. Experimental dataset details.

| No. | Name | Row | Column | Known Entries | Density |
|---|---|---|---|---|---|
| D1 | MovieLens10M [37] | 71,567 | 10,681 | 10,000,054 | 1.31% |
| D2 | MovieLens20M [37] | 138,493 | 26,744 | 20,000,263 | 0.54% |
| D3 | Douban [38] | 129,490 | 58,541 | 16,830,839 | 0.22% |
| D4 | Jester [38] | 124,113 | 150 | 5,865,235 | 31% |

In order to acquire objective results, all known entry set of each HDI date is split into 10 disjoint and equally-sized subsets randomly, and choose seven subsets as the training set, one set as the validation set, remaining two sets as the testing set. And above process is repeated five times for five-fold cross-validation. The conditions of termination are: a) the number of training iterations exceeds preset threshold, i.e., 1000; b) the model converges, i.e., the error difference between two consecutive iterations is smaller than $10^{-5}$.

Moreover, there are several settings in our experiment for more compelling results:

a) The LF matrices $X$ and $Y$ are initialized with the same randomly generated arrays for all compared LFA models;
b) The regularization coefficients are set as 0.05 for avoiding overfitting, and the learning rate is set as $\eta=0.04$ for SGD-based LFA model [2];
c) For balance the computational efficiency and representative ability of learning models, the dimension of LF space $f$ is set as $f=20$ uniformly [8].

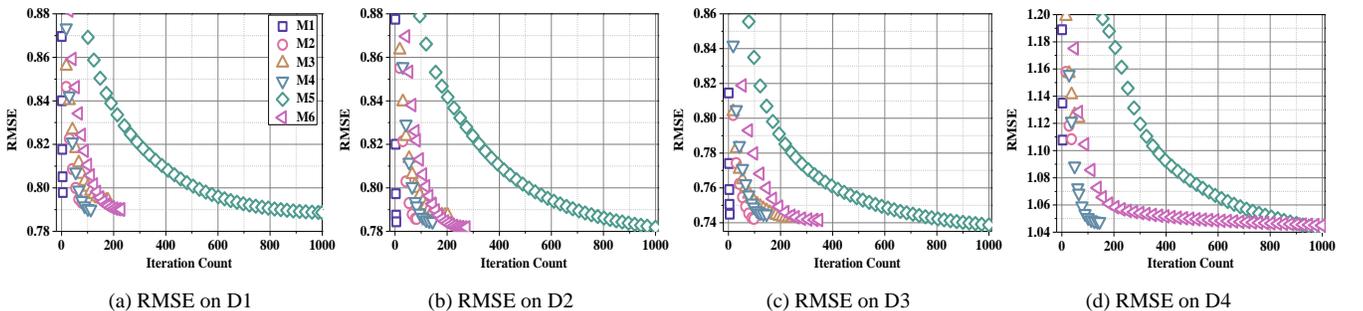

(a) RMSE on D1  (b) RMSE on D2  (c) RMSE on D3  (d) RMSE on D4

Fig.1. Training curves of M1-M6 on D1-D4 in RMSE.

## B. Comparison against State-of-the-art Models

In this section, we give the comparative experimental results of NPALF-based LFA model with several state-of-art LFA models. Table II records the details of all compared models. Table III records the detailed performance of M1-M6 on D1-D4. Figs. 1 depict the training curves of M1-M6. From those results, we have following findings:

a) **M1's prediction efficiency is significantly higher than other models.** For instance, as depicted in Table III and Fig. 1, M1, i.e., a proposed NPALF model, just consumes 109.5 seconds to converge in RMSE on D2, compared with others, M1's computational efficiency is improved about 26% (i.e., ($Cost_{high}$-$Cost_{low}$)/$Cost_{high}$) than M2's 149.4 seconds, 64% than M3's 305.2 seconds, 93% than M4's 1645.8 seconds, 98% than M5's 6250.8 seconds, and 91% than M6's 1265.6 seconds. However, we find that M1's total time-consumption is higher than M2 in terms of RMSE on D1, this means that the NPALF model is data-dependent. Same outcomes are also seen on other datasets as depicted in Table III and Fig.1.

b) **M1 also achieve quite satisfactory prediction accuracy for missing data of an HDI matrix when compared with other models.** According to Table III, M1 implement the lowest prediction error on three testing cases out of four in total, and small gap in predictions in remaining test cases. For instance, M2's RMSE is 0.7837 on D1, which is about 0.16% lower than 0.7850 by M2, 0.36% lower than 0.7866 by M3, 1.8% lower than 0.7981 by M4, 3.1% lower than 0.8094 by M5, and 2.9% lower than 0.8072 by M6. Moreover, similar results are proved on the other testing cases as shown in Table III.

TABLE II. Details of compared models.

| Model | Name | Description |
|---|---|---|
| M1 | NPALF | The proposed NPALF-based LFA model. |
| M2 | PLF | A standard PID-incorporated SGD-based LFA model [38]. |
| M3 | SGD-LFA | A standard SGD-based LFA model [50]. |
| M4 | Adam | An Adam-based LFA model [50]. |
| M5 | AdaDeleta | An Ada-Delta-based LFA model [50]. |
| M6 | RMSprop | An RMSprop-based LFA model [50]. |

TABLE III. Lowest RMSE and their corresponding total time cost (Secs).

| Case | | M1 | M2 | M3 | M4 | M5 | M6 |
|---|---|---|---|---|---|---|---|
| **D1** | RMSE: | **0.7914±5E-3** | 0.7929±3E-4 | 0.7939±5E-4 | 0.8045±6E-5 | 0.8155±6E-4 | 0.8166±6E-3 |
| | Time: | 77.7±1.0 | **71.4±1.5** | 134.4±1.2 | 615.5±2.3 | 3588.4±3.8 | 509.8±2.4 |
| **D2** | RMSE: | **0.7837±5E-4** | 0.7850±5E-4 | 0.7866±5E-3 | 0.7981±6E-3 | 0.8094±6E-4 | 0.8072±6E-3 |
| | Time: | **109.5±1.3** | 149.4±1.8 | 305.2±1.5 | 1645.8±3.2 | 6250.8±3.9 | 1265.6±2.3 |
| **D3** | RMSE: | **0.7245±4E-4** | 0.7252±5E-4 | 0.7257±5E-4 | 0.7390±6E-4 | 0.7446±6E-4 | 0.7452±6E-4 |
| | Time: | **118.3±1.8** | 150.1±1.6 | 336.8±1.3 | 1402.0±3.1 | 5812.2±3.6 | 1350.4±2.3 |
| **D4** | RMSE: | 1.1056±3E-3 | 1.1047±4E-4 | 1.1220±4E-4 | 1.0530±6E-4 | 1.0453±6E-4 | **1.0427±5E-3** |
| | Time: | **4.3±1.3** | 5.2±0.7 | 8.0±0.6 | 146.4±2.1 | 563.2±3.8 | 390.8±3.0 |

## V. CONCLUSIONS

This paper innovatively proposes an NPALF model, which incorporates the NPID controller into an SGD-based LF model to achieve fast convergence rate and satisfactory prediction accuracy for missing data of an HDI matrix. Meanwhile, we adopt PSO method into NPALF-based LFA model for implementing all parameter adaptation. Experimental results show that it improves prediction accuracy and computational efficiency significantly.

## REFERENCES

[1] X. Luo, Z-G. Liu, S. Li, M-S. Shang, and Z-D. Wang, "A Fast Non-negative Latent Factor Model based on Generalized Momentum Method," *IEEE Trans. on System Man Cybernetics: Systems*, vol. 51, no. 1, pp. 610-620, 2021.

[2] X. Luo, M-C. Zhou, S. Li, Y-N. Xia, Z-H. You, Q-S. Zhu, and H. Leung, "Incorporation of efficient second-order solvers into latent factor models for accurate prediction of missing QoS data," *IEEE Trans. on Cybernetics,* vol. 48, no.4, pp. 1216-1228, 2018.

[3] X. Luo, Ye, Yuan, S-L. Chen, N-Y. Zeng, and Z-D. Wang, "Position-Transitional Particle Swarm Optimization-incorporated Latent Factor Analysis," *IEEE Trans. on Knowledge and Data Engineering*, DOI: 10.1109/TKDE.2020.3033324.

[4] D. Wu, X. Luo, M-S. Shang, Y. He, G-Y. Wang, and X-D. Wu, "A Data-Characteristic-Aware Latent Factor Model for Web Services QoS Prediction," *IEEE Trans. on Knowledge and Data Engineering*, DOI: 10.1109/TKDE.2020.3014302, 2020.

[5] D. Wu, and X. Luo, "Robust Latent Factor Analysis for Precise Representation of High-dimensional and Sparse Data," *IEEE/CAA Journal of Automatica Sinica*, vol. 8, no. 4, pp. 796-805, 2021.

[6] X. Luo, H. Wu, and Z-C Li, "NeuLFT: A Novel Approach to Nonlinear Canonical Polyadic Decomposition on High-Dimensional Incomplete Tensors," *IEEE Trans. on Knowledge and Data Engineering*, DOI: 10.1109/TKDE.2022.3176466, 2022.

[7] X. Luo, Y. Zhou, Z-G. Liu, and M-C. Zhou, "Fast and Accurate Non-negative Latent Factor Analysis on High-dimensional and Sparse Matrices in Recommender Systems," *IEEE Trans. on Knowledge and Data Engineering*, DOI: 10.1109/TKDE.2021.3125252, 2021.

[8] X. Luo, H. Wu, Z. Wang, J-J. Wang, and D-Y. Meng, "A Novel Approach to Large-Scale Dynamically Weighted Directed Network Representation," *IEEE Trans. on Pattern Analysis and Machine Intelligence*, DOI: 10.1109/TPAMI.2021.3132503, 2021.

[9] X. Luo, H. Wu, H. Yuan, M-C. Zhou, "Temporal pattern-aware QoS prediction via biased non-negative latent factorization of tensors," *IEEE trans. on cybernetics*, vol. 50, no. 5, pp. 1798-1809, 2019.

[10] Z-B. Li, S. Li, and X. Luo, " An overview of calibration technology of industrial robots," *IEEE/CAA Journal of Automatica Sinica*, vol. 8, no. 1, pp. 23-36, 2021.

[11] X. Luo, Z. Liu, L. Jin, Y. Zhou, M-C. Zhou, " Symmetric nonnegative matrix factorization-based community detection models and their convergence analysis," *IEEE Trans. on Neural Networks and Learning Systems,* vol. 33, no. 3, pp. 1203-1215, 2021.

[12] L. Hu, S. Yang, X. Luo, and M-C. Zhou, " An algorithm of inductively identifying clusters from attributed graphs", *IEEE Trans. on Big Data*, vol.8, no. 2, pp. 523-534, 2020.

[13] L. Hu, X. Pan, Z. Tan, amd X. Luo, "A fast fuzzy clustering algorithm for complex networks via a generalized momentum method," *IEEE Trans. on Fuzzy Systems,* DOI: 10.1109/TFUZZ.2021.3117442, 2021.

[14] L. Hu, S. Yang, X. Luo, H. Yuan, K. Sedraoui, and M-C. Zhou, " A distributed framework for large-scale protein-protein interaction data analysis and prediction using mapreduce," *IEEE/CAA Journal of Automatica Sinica*, vol. 9, no. 1, pp. 160-172, 2021.

[15] M. S. Shang, Y. Yuan, X. Luo, and M. C. Zhou, "An *α-β*-divergence-generalized Recommender for Highly-accurate Predictions of Missing User Preferences," *IEEE Trans. on Cybernetics*, DOI: 10.1109/TCYB.2020.3026425.

[16] Q. Shi, H-K. Lam, and C-B. Xuan, "Adaptive neuro-fuzzy PID controller based on twin delayed deep deterministic policy gradient algorithm," *neurocomputing*, vol. 402, pp. 183-194, 2020.

[17] J. Cuellar, A. Y; de Jesus, R. Troncoso, and M. Velazquez, "PID-controller tuning optimization with genetic algorithms in servo system," *Int. J. Adv. Robot. Syst.*, vol. 10, no. 9, pp. 324, 2013.

[18] J-D. Zhang, Z-J. Li, W-B. Meng, "The PID algorithm for multi-vehicles cruise control in platoon," *Journal of Information and Computational Science*, vol. 12, no. 10, pp. 3899-3906, 2015.

[19] M. Li, Y. Zhang, D-Z. You, "Design of fuzzy PID stepping motor controller based on particle swarm optimization." in *Proc. of the 3rd Conf. on Mechanical Engineering and Intelligent Manufacturing*, pp. 449-453, 2020.

[20] X. Y. Shi, Q. He, X. Luo, Y. N. Bai, and M. S. Shang, "Large-scale and Scalable Latent Factor Analysis via Distributed Alternative Stochastic Gradient Descent for Recommender Systems," *IEEE Trans. on Big Data*, DOI: 10.1109/TBDATA.2020.2973141, 2020.

[21] H. Wu, X. Luo, and M. C. Zhou, "Advancing Non-negative Latent Factorization of Tensors with Diversified Regularizations," *IEEE Trans. on Services Computing*, DOI: 10.1109/TSC.2020.2988760, 2020.

[22] A-L. Salih, M. Moghavvemi, H-A. Mohamed, and K-S. Gaeid, "Modelling and pid controller design for a quadrotor unmanned air vehicle," in *Proc. of the IEEE Int. Conf. on Automation, Quality and Testing, Robotics*, pp. 1-5, 2010.

[23] G-M. Riduwan, A-M. Ashraf, I-R. Mohd and T. Raja, "A Multiple-node Hormone Regulation of Neuroendocrine-PID(MnHR-NEPID) Control for Nonlinear MIMO Systems," *IETE journal of research*, 2020.

[24] D. Wu, M. S. Shang, X. Luo, and Z. D. Wang, "An $L_1$-and-$L_2$-norm-oriented Latent Factor Model for Recommender Systems, *IEEE Trans. on Neural Networks and Learning Systems*, DOI: 10.1109/TNNLS.2021.3071392, 2021.

[25] D. Wu, Y. He, X. Luo, and M. C. Zhou, "A Latent Factor Analysis-based Approach to Online Sparse Streaming Feature Selection," *IEEE Trans. on System Man Cybernetics: Systems*, DOI: 10.1109/TSMC.2021.3096065, 2021.

[26] P. Massa, and P. Avesani, "Trust-aware recommender systems," in *Proc. of the 1st ACM Conf. on Recommender Systems*, pp. 17-24, 2007.

[27] X. Luo, Y. Yuan, M. C. Zhou, Z. G. Liu, and M. S. Shang, "Non-negative Latent Factor Model based on *β*-divergence for Recommender Systems," *IEEE Trans. on System, Man, and Cybernetics: Systems*, vol. 51, no. 8, pp. 4612-4623, 2021.

[28] X. Luo, Z. G. Liu, S. Li, M. S. Shang, and Z. D. Wang, "A Fast Non-negative Latent Factor Model based on Generalized Momentum Method," *IEEE Trans. on System, Man, and Cybernetics: Systems*, vol. 51, no. 1, pp. 610-620, 2021.

[29] Z. G. Liu, X. Luo, and Z. D. Wang, "Convergence Analysis of Single Latent Factor-Dependent, Nonnegative, and Multiplicative Update-Based Nonnegative Latent Factor Models," *IEEE Trans. on Neural Networks and Learning Systems*, vol. 32, no. 4, pp. 1737-1749, 2021.

[30] R-Z. Song, and L. Zhu, "Optimal fixed-point tracking control for discrete-time nonlinear systems via ADP," *IEEE/CAA J. Autom. Sinica*, vol. 6, no. 3, pp. 657-666, 2019.

[31] Y. Yuan, Q. He, X. Luo, and M. S. Shang, "A multilayered-and-randomized latent factor model for high-dimensional and sparse matrices," *IEEE Trans. on Big Data*, DOI: 10.1109/TBDATA.2020.2988778, 2020.

[32] X. Luo, Y. Zhou, Z. G. Liu, L. Hu, and M. C. Zhou, "Generalized Nesterov's Acceleration-incorporated, Non-negative and Adaptive Latent Factor Analysis," *IEEE Trans. on Services Computing*, DOI: 10.1109/TSC.2021.3069108, 2021.

[33] X. Luo, Z. D. Wang, and M. S. Shang, "An Instance-frequency-weighted Regularization Scheme for Non-negative Latent Factor Analysis on High Dimensional and Sparse Data," *IEEE Trans. on System Man Cybernetics: Systems*, vol. 51, no. 6, pp. 3522-3532, 2021.

[34] L. Hu, X. H. Yuan, X. Liu, S. W. Xiong, and X. Luo, "Efficiently Detecting Protein Complexes from Protein Interaction Networks via Alternating Direction Method of Multipliers," *IEEE/ACM Trans. on Computational Biology and Bioinformatics*, vol. 16, no. 6, pp. 1922-1935, 2019.

[35] L. Fan, E-M. Joo, "Design for auto-tuning PID controller based on genetic algorithms," in *Proc. of the 4th IEEE Conf. on Industrial Electronics and Applications*, pp. 1924-1928, 2009.

[36] X. Luo, Z. G. Liu, M. S. Shang, J. G. Lou, and M. C. Zhou, "Highly-Accurate Community Detection via Pointwise Mutual Information-Incorporated Symmetric Non-negative Matrix Factorization," *IEEE Trans. on Network Science and Engineering*, vol. 8, no. 1, pp. 463-476, 2021.

[37] J-L. Li, Y. Yuan, "A Nonlinear Proportional Integral Derivative-Incorporated Stochastic Gradient Descent-based Latent Factor Model," *IEEE Int. Conf. on Systems, Man and Cybernetics,* DOI: 10.1109/SMC42975.2020.9283344.

[38] J-L. Li, X-K. Wu, Y. Yuan, Y-J. Wu, K-K. M, Y. Zhou, "Accelerated Latent Factor Analysis for Recommender Systems via PID Controller," in *Proc. of the 31th ICNSC*, DOI: 10.1109/ICNSC48988.2020.9238055, 2020.

[39] L. Hu, P. W. Hu, X. Y. Yuan, X. Luo, and Z. H. You, "Incorporating the Coevolving Information of Substrates in Predicting HIV-1 Protease Cleavage Sites," *IEEE/ACM Trans. on Computational Biology and Bioinformatics*, vol. 17, no. 6, pp. 2017-2028, 2020.

[40] S. Zhang, L. Yao, B. Wu, X. Xu, X. Zhang, and L. Zhu, "Unraveling Metric Vector Spaces With Factorization for Recommendation," *IEEE Trans. on Industrial Informatics*, vol. 16, no. 2, pp. 732-742, 2020.

[41] X. Luo, W. Qin, A. Dong, K. Sedraoui, and M. C. Zhou, "Efficient and High-quality Recommendations via Momentum-incorporated Parallel Stochastic Gradient Descent-based Learning," *IEEE/CAA Journal of Automatica Sinica*, vol. 8, no. 2, pp. 402-411, 2021.

[42] X. Luo, Y. Yuan, S. L. Chen, N. Y. Zeng, and Z. D. Wang, "Position-Transitional Particle Swarm Optimization-Incorporated Latent Factor Analysis," IEEE Trans. on Knowledge and Data Engineering, DOI: 10.1109/TKDE.2020.3033324, 2020.

[43] D. Wu, Q. He, X. Luo, M. S. Shang, Y. He, and G. Y. Wang, "A posterior-neighborhood-regularized latent factor model for highly accurate web service QoS prediction," *IEEE Trans. on Services Computing*, DOI: 10.1109/TSC.2019.2961895, 2019.

[44] X. Luo, M. C. Zhou, Y. N. Xia, and Q. S. Zhu, "An efficient non-negative matrix-factorization-based approach to collaborative filtering for recommender systems," *IEEE Trans. on IndustrialInformatics*, vol. 10, no. 2,pp. 1273‑1284, 2014.

[45] L. Bottou, "Large-Scale Machine Learning with Stochastic Gradient Descent," *Physica-Verlag HD*, 2010.

[46] R. C. Eberhart and Y. H. Shi, "Particle swarm optimization: developments, applications and resources," in *Proc. of the Congress on Evolutionary Computation*, pp. 81-86, 2001.

[47] S-D. Hanwate, Y-V. Hole, "Optimal PID design for Load frequency control using QRAWCP approach," *IFAC-PapersOnLine*, vol. 51, no. 4, pp. 651-656, 2018.

[48] P. Zhao, J-J. Chen, Y. Song, X. Tao, T-J. Xu and T. Mei, "Design of a control system for an autonomous vehicle based on adaptive-pid," *International Journal of Advanced Robotic Systems*, vol. 9, no. 2, pp. 44, 2012.

[49] C. Jiang, Y. Ma, C. Wang, "PID controller parameters optimization of hydro-turbine governing systems using deterministic-chaotic-mutation evolutionary programming," *Energy Convers. Manag*. vol. 47, no. 9-10, pp. 1222-1230, 2006.

[50] X. Luo, D-X Wang, M-C Zhou, and H. Yuan, "Latent Factor-Based Recommenders Relying on Extended Stochastic Gradient Descent Algorithms," *IEEE Trans. on Systems, Man, and Cybernetics: Systems*, DOI. 10.1109/TSMC, 2018.